\documentclass{article}
\usepackage[utf8]{inputenc}
\usepackage{graphicx}
\usepackage{multirow}
\usepackage{url}
\usepackage{hyperref}
\usepackage[table]{xcolor}
\usepackage{amsmath}
\usepackage{amssymb}
\usepackage{diagbox}
\usepackage{hhline}
\usepackage{arxiv}

\title{ASAP: Adaptive Transmission Scheme for Online Processing of Event-based Algorithms}

\author{
    \href{https://orcid.org/0000-0002-4435-5466}{\includegraphics[scale=0.06]{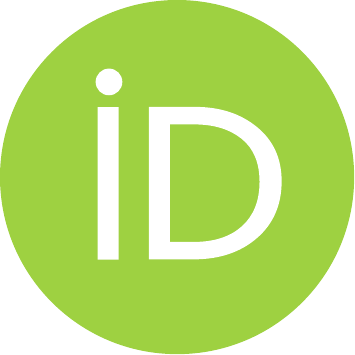}\hspace{1mm}Raul Tapia}\\
	GRVC Robotics Lab.\\
	Universidad de Sevilla\\
	\texttt{raultapia@us.es}\\
	\And
	\href{https://orcid.org/0000-0001-9431-7831}{\includegraphics[scale=0.06]{orcid.pdf}\hspace{1mm}José Ramiro Martínez-de Dios} \\
	GRVC Robotics Lab.\\
	Universidad de Sevilla\\
	\texttt{jdedios@us.es}\\
	\And
	\href{https://orcid.org/0000-0002-2285-2605}{\includegraphics[scale=0.06]{orcid.pdf}\hspace{1mm}Augusto Gómez Eguíluz} \\
	GRVC Robotics Lab.\\
	Universidad de Sevilla\\
	\texttt{ageguiluz@us.es}\\
	\And
	\href{https://orcid.org/0000-0003-2155-2472}{\includegraphics[scale=0.06]{orcid.pdf}\hspace{1mm}Anibal Ollero} \\
	GRVC Robotics Lab.\\
	Universidad de Sevilla\\
	\texttt{aollero@us.es}\\
}

\paperurl{https://raultapia.com/preprints/0002/redirect}
\journal{Autonomous Robots}
\editorial{Springer}

\begin{document}

\maketitle

\begin{abstract}
Online event-based perception techniques on board robots navigating in complex, unstructured, and dynamic environments can suffer unpredictable changes in the incoming event rates and their processing times, which can cause computational overflow or loss of responsiveness. This paper presents ASAP: a novel event handling framework that dynamically adapts the transmission of events to the processing algorithm, keeping the system responsiveness and preventing overflows. ASAP is composed of two adaptive mechanisms. The first one prevents event processing overflows by discarding an adaptive percentage of the incoming events. The second mechanism dynamically adapts the size of the \textit{event packages} to reduce the delay between event generation and processing. ASAP has guaranteed convergence and is flexible to the processing algorithm. It has been validated on board a quadrotor and an ornithopter robot in challenging conditions.
\end{abstract}

\newcommand{\citep}[1]{\cite{#1}}

\section{Introduction}
\label{sec:intro}
In the last years event cameras have attracted increasing interest in the robotics community. Event cameras are neuromorphic sensors that capture asynchronous illumination changes at pixel level with $\mu$s resolution. They are insensitive to motion blur and have a wide dynamic range, providing high robustness to lighting conditions. A good number of successful event processing techniques have been proposed evidencing their capabilities, see e.g. \citep{gallego2020event}.

Most existing event handling frameworks deliver the events generated by the camera to the event processing algorithm in \textit{event packages} --i.e., structures in which several events are buffered to reduce the transmission cost. Packaging using a fixed number of events --i.e., fixed package size-- or a fixed triggering rate are the most common approaches. Event delivery has a critical impact on the processing algorithm performance. For instance, in cases with low event generation rates, using large packages can generate unnecessary delay, whereas short packages would enable better exploiting the asynchronous nature of event cameras. On the other hand, in cases with high event generation rates, using short packages would require devoting significant time to transmit and receive the events, increasing the risk of processing bottlenecks and saturation. When using event cameras on board robots in complex environments, changes in lighting conditions, type of scene, or robot motion can induce strong unpredictable changes in the event rate and in the processing algorithm efficiency. Static delivery is configured for the expected event rate and algorithm processing time, and cannot adapt to these changes, often resulting in processing responsiveness loss or computational overflow.

\begin{figure}[ht]
\begin{center}
\includegraphics[width=0.5\linewidth]{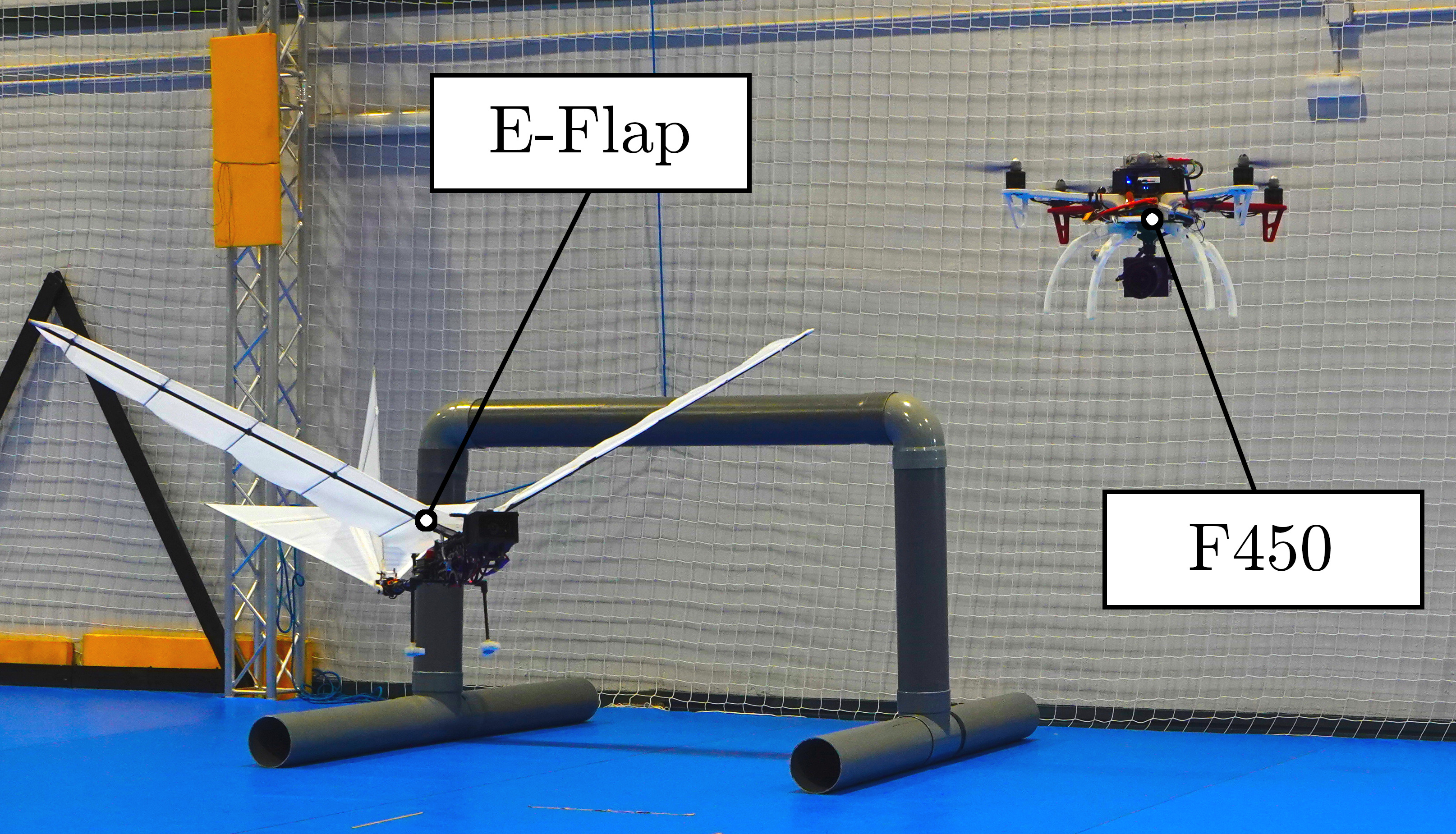}
\caption{Picture from ASAP validation experiments on board a DJI Flamewheel F450 and the GRIFFIN \textit{E-Flap} ornithopter.}
\label{fig:robots}
\end{center}
\end{figure}

This paper presents ASAP, an event handling framework that dynamically adapts event delivery to enhance processing responsiveness and prevent computational overflows. It absorbs the variations in the event rates and in the algorithm event processing time through two adaptive feedback-based mechanisms that run simultaneously and have a synergistic behaviour in which they help one another in the adaptation. Any event-based algorithm can be fed with ASAP requiring only minor implementation modifications. The proposed event handling scheme is flexible to the processing algorithm, its computational complexity, and its implementation, and also offers the possibility of packaging with fixed time windows and fixed number of events. The main contributions of this paper are:
\begin{enumerate}
    \item an adaptive event handling scheme that enables responsive event processing with no overflows;
    \item its experimental validation and evaluation on board a quadrotor and an ornithopter robot, see Fig. \ref{fig:robots};
    \item ASAP code\footnote{\url{https://raultapia.com/repositories/asap}} to contribute to the robotics, computer vision, and neuromorphic sensing communities.
\end{enumerate}

This work is inspired by our previous workshop paper \citep{tapia2020asap}. The main novelties are: a) deep improvements in both adaptive mechanisms to take into account the temporal cost fed back from the event processing algorithm; b) improvement in the event packaging mechanism, which can now be tuned using closed analytical expressions; c) mathematical demonstration of its convergence; and d) new and detailed experimental validation including the implementation and evaluation in two aerial robots in challenging conditions.

The structure of the paper is as follows. The main related work is summarized in Section \ref{sec:soa}. The proposed scheme and its two mechanisms are presented in Sections \ref{sec:general}, \ref{sec:bucle2}, and \ref{sec:bucle1}, respectively. The convergence demonstration is described in Section \ref{sec:conver}. The experimental evaluation and validation are described in Section \ref{sec:results}. Section \ref{sec:conclusions} summarizes the conclusions and future work.

\section{Related Work}
\label{sec:soa}
Existing event-based methods can be classified into: \textit{event images}, \textit{event-by-event}, and \textit{hybrid} algorithms. The most widely used approach in robotics relies on processing \textit{event images}, created by accumulating the events received. A method for detection and tracking using a Micro Aerial Vehicle (MAV) was developed in \citep{mitrokhin2018event}. It compensated for the global motion of the robot using a model of the affine transformation between two consecutive \textit{event images}. Work \citep{falanga2020dynamic} presented a reactive obstacle avoidance method for quadrotors using \textit{event images} from a stereo set-up. All these works provided valid solutions to some existing and novel problems. However, processing \textit{event images} does not fully exploit the asynchronous behaviour of event cameras.

\textit{Event-by-event} methods process events one by one without grouping, exploiting the asynchronous nature of the event cameras. A number of solutions have been developed for feature detection \citep{vasco2016fast} \citep{mueggler2017fast} \citep{li2019fa}, tracking \citep{alzugaray2018asynchronous}, and clustering \citep{barranco2018real}, among others. Although these methods are designed to process events one-by-one, few of them have focused on mobile robot applications, where computational restrictions can be relevant. An \textit{event-by-event} pose tracking method for Unmanned Aerial Vehicles (UAVs) was developed in \citep{mueggler2014event}. The method was validated during high-speed maneuvers and reached accurate pose estimations. An event-based surveillance system using UAVs was presented in \citep{rodriguez2020asynchronous}. The authors proposed a fully \textit{event-by-event} scheme that combines feature detection, tracking, and clustering.

Although \textit{event-by-event} processing overcomes some of the limitations of \textit{event images}, events are noisy and processing one-by-one results in high computational cost. \textit{Hybrid} approaches combine the advantages of \textit{event-by-event} and \textit{event image} processing. A \textit{hybrid} motion detector method was implemented in \citep{vasco2017independent} by learning the motion model of corner features when no object was present. The work in \citep{gomez2020asynchronous} presented a bioinspired approach to time-to-contact maneuvers for multirotors.

Although most of the above processing methods provide valid solutions to robotic perception, the online implementation of event-based vision systems on real robots requires mechanisms capable of efficiently managing the computational resources. However, few works have addressed the efficient delivery of events between event cameras and event processing algorithms. Existing event handling frameworks can be classified according to the manner in which they feed the event processing algorithms. The use of \textit{event packages} is the most widespread trend. The event-based framework included into the YARP (\textit{Yet Another Robot Platform}) architecture \citep{glover2018event} or the RPG ROS DVS driver --used, for instance, in \citep{mueggler2014event}, \citep{lichtsteiner2008latency} or \citep{brandli2014latency}-- rely on the use of \textit{event packages} to feed the processing algorithms. Packages can be built up by accumulating a fixed number of events or by accumulating events during a fixed period of time. Although this kind of event handling prevents the algorithm's communication overhead to some extent, they are not efficient in cases when the processing of one \textit{event package} has been completed before the next package arrives. On the other hand, \textit{event-by-event} transmission is an efficient solution when algorithms can process events faster than they are received. The work in \citep{marcireau2019sepia} presented a framework for \textit{event-by-event} handling that consists of three modules: an I/O library (called \textit{Sepia}), a computation toolbox (\textit{Tarsier}), and a visualization tool (\textit{Chameleon}). Their framework suppresses the use of buffers between the modules and limits the use of event buffers to the reception of events from the camera. Although \textit{event-by-event} frameworks provide a lower communication delay than solutions based on \textit{event packages}, its use is restricted to algorithms that complete the processing of an event before the reception of the following. Otherwise, high event rates can cause bottlenecks and, eventually, saturation.

Online event processing on board real robots requires effective management of the computational load. Some works have started to explore how to adapt the incoming event stream from the event camera. Since \citep{berner2007address}, removing events that cannot be delivered in time has been a solution adopted by the event camera logic interface. Other works such as \citep{rodriguez2020asynchronous} or \citep{tapia2020asap} used the idea of randomly discarding a percentage of the incoming events to reduce the number of events to be processed. A feedback control scheme was presented in \citep{delbruck2021feedback} for the adaptive setting of the internal parameters of the DVS sensor. Their method regulates the event rate by controlling the threshold and refractory period changing the internal bias currents, and also regulates noise using bandwidth control. Regarding adaptive packaging, the work in \citep{glover2018controlled} adjusts \textit{event package} size by finding the error between the event rate produced by the camera and the event rate that can be processed.

The method proposed in this paper focuses on event delivery under strong event rate changes such as those that can be found in aerial robotics in complex and dynamic environments with low-resource onboard processing capacity. It dynamically adapts event delivery by using two closed-loop mechanisms with synergistic behaviours that help one another in the adaptation. To the best of the authors' knowledge, it is the first event adaptive handling framework validated on board aerial robots performing aggressive maneuvers.

\section{General Description}
\label{sec:general}
Consider an event camera, which events are delivered for processing at an event processing algorithm. Each event is represented as $e_i$ and includes the timestamp at which it was generated, the event pixel coordinates, and the event polarity. Events are transmitted from the pixel array to the output bus by using Address Event Representation (AER). In many applications, the event rates registered by the event camera may suffer unpredictable changes. For instance, that is the case in many robotics applications in which ground or aerial robots equipped with onboard event cameras navigate in complex and dynamic environments. Although static \textit{event packages}, using either a fixed number of events or a fixed triggering rate, is the most widely approach, it can cause processing saturation and loss of responsiveness in cases with strong event rate changes. If the event generation rate is lower than expected, static packaging can cause unnecessary delays between event generation and processing. If it is higher, it can cause processing bottlenecks and overflows, since the algorithm may receive a new \textit{event package} before completing the processing of the previous one. 

Two metrics are interesting to analyze the transmission of an \textit{event package} $p_k$: the \textit{delivery time} $\tau_k$ and the package \textit{building time} $\pi_k$. First, $\tau_k$ is defined as the difference between the time at which $p_k$ is sent to the processing algorithm and the time the processing of $p_k$ starts --including a possible package buffer. $\pi_k$ is defined as the difference between the timestamps of the oldest and newest event in $p_k$. In addition, the term \textit{latency} could be applied to different time intervals regarding generation, packaging, buffering, and processing. As a convention, this work defines the \textit{latency} $\lambda_k$ of $p_k$ as the time between the generation of the oldest event in $p_k$ and the start of the processing of $p_k$, i.e. $\pi_k + \tau_k$.

\begin{figure}[ht]
\centering
\includegraphics[width=0.75\linewidth]{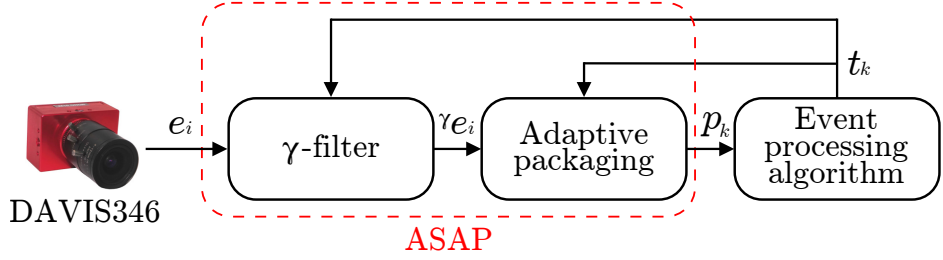}
\caption{General scheme of ASAP.}
\label{fig:general_scheme}
\end{figure}

The objective of ASAP is to dynamically adapt the transmission of events from the camera to the processing algorithm to keep its responsiveness and prevent computational overflow by feeding the algorithm with a new package \textit{as soon as possible}, i.e. as soon as it has finished the processing of the previous package. The scheme of ASAP is shown in Fig. \ref{fig:general_scheme}. It includes two closed-loop adaptive mechanisms, both use $t_{k-1}$, the time required by the algorithm to process the events from the last package $p_{k-1}$. The first mechanism, based on the adaptive packaging module, enhances the algorithm responsiveness by dynamically adapting the size of the \textit{event packages}. It uses $t_{k-1}$ to select $s_k$, the size of the next \textit{event package}. Adapting the \textit{event package} size can be insufficient for preventing processing overflows in case of high event rates. The second mechanism, based on module $\gamma$-filter, prevents overflows by randomly removing an adaptive percentage of events that will not be packaged or processed. In previous work \citep{rodriguez2020asynchronous} we noticed that in cases with high event rates, removing up to a maximum percentage of events had a low impact on the algorithm performance. $\gamma$-filter measures the current event generation rate and uses it to dynamically select the percentage of events to be removed. Event removal percentage is minimum at low event rates and increases with the event rate. 

Both closed-loop mechanisms are executed simultaneously and interact exhibiting a joint synergistic behaviour that enables fast adaptation. For instance, if $t_{k-1}$ has increased, the adaptive packaging module reacts by increasing $s_k$. This will increase the \textit{latency} $\lambda_k$, losing responsiveness. Simultaneously, due to the higher $t_{k-1}$, $\gamma$-filter reacts by increasing the event removal percentage, reducing $\lambda_k$ and improving responsiveness. Also, in the opposite case, if $t_{k-1}$ has decreased, adaptive packaging reduces $s_k$, and $\gamma$-filter reduces the event removal percentage. The notation of the main variables used in ASAP is shown in Table \ref{table:notation}.

\begin{table}[ht]
\begin{center}
\setlength\arrayrulewidth{1pt}
\begin{tabular}{|c|c|}
\hline
\cellcolor{gray!25}\textbf{Variable} & \cellcolor{gray!25}\textbf{Description}\\ \hline
$e_i$ & $i$-th event of the incoming event stream \\ \hline
${^\gamma}e_i$ & $i$-th event of the event stream after $\gamma$-filter  \\ \hline
$r_i$ & Incoming event rate at the time when $e_i$ arrives \\ \hline
${^\gamma}r_i$ & $\gamma$-filtered event rate at the time when $e_i$ arrives \\ \hline
$p_k$ & $k$-th \textit{event package} generated by ASAP\\ \hline
$s_k$ & Number of events in the \textit{event package} $p_k$ \\ \hline
$t_k$ & Time required by the algorithm to process $p_k$ \\ \hline
\multirow{2}{*}{$\tau_k$} & Temporal difference between $p_k$ is sent to \\
                          & the algorithm and processing of $p_k$ starts \\ \hline
\multirow{2}{*}{$\pi_k$} & Temporal difference between the timestamps \\
                         & of the oldest and newest event in $p_k$ \\ \hline
$\lambda_k$ & \textit{Latency} of $p_k$, computed as $\tau_k + \pi_k$ \\ \hline
\end{tabular}
\end{center}
\caption{Summary of the notation used.}
\label{table:notation}
\end{table}

\section{Adaptive Event Packaging}
\label{sec:bucle2}
This mechanism adapts the size of the \textit{event packages} delivered to the processing algorithm to reduce the delay between event generation and processing. It uses $t_{k-1}$, the time required by the algorithm to process the events in the last package $p_{k-1}$, to select $s_k$, the number of events to be included in the next package $p_k$. Package $p_k$ is created by buffering events until the size of $p_k$ is $\geq s_k$. Then, $p_k$ is delivered for processing.

There is a dependence between the number of events in the package and the time required to process the packet. On the one hand, smaller packages will require lower time to be processed. On the other hand, the processing of events delivered in smaller packages is less efficient due to the time required to transmit and receive the events. Adaptive packaging selects $s_k$ according to $t_{k-1}$ using the following generic function: 
\begin{equation}
s_k = A\Phi(t_{k-1}) + B,
\label{eq:s_generic}
\end{equation}
where $A, B \in \mathbb{R}$ enable adjusting the ranges of $s_k$ and $t_{k-1}$, namely $[s^{\mathrm{min}}, s^{\mathrm{max}}]$ and $[t^{\mathrm{min}},t^{\mathrm{max}}]$. $A$ and $B$ should be adjusted such that $s^{\mathrm{min}} = A\Phi(t^{\mathrm{min}}) + B$ and $s^{\mathrm{max}} = A\Phi(t^{\mathrm{max}}) + B$. Hence, $A$ and $B$ should satisfy:
\begin{equation}
A = \frac{s^{\mathrm{max}} - s^{\mathrm{min}}}{\Phi(t^{\mathrm{max}}) - \Phi(t^{\mathrm{min}})},
\label{eq:a}
\end{equation}
\begin{equation}
B = s^{\mathrm{max}} - A \Phi(t^{\mathrm{max}}).
\label{eq:b}
\end{equation}
$\Phi$ is a continuous function that relates $t_{k-1}$ and $s_k$. ASAP can use any function $\Phi$ that satisfies the following conditions:

\begin{enumerate}
    \item Every value of $t_{k-1} \in [t^{\mathrm{min}}, t^{\mathrm{max}}]$ must have an associated value of $s_k$. Hence, $\Phi$ should be defined $\forall t_{k-1} \in [t^{\mathrm{min}}, t^{\mathrm{max}}]$.
    \item Larger package sizes should be assigned to higher processing times. Hence, $\Phi$ should be monotonically increasing, i.e. $\frac{\partial \Phi(t_{k-1})}{\partial t_{k-1}} > 0, \forall t_{k-1} \in [t^{\mathrm{min}}, t^{\mathrm{max}}]$.
    \item Event packaging should have ranges with different sensitivity of $s_k$ against variations in $t_{k-1}$. Hence, $\Phi$ should have at least one inflection point $t^{\mathrm{flex}}$, i.e. $\exists t^{\mathrm{flex}} : \frac{\partial^2 \Phi(t_{k-1})}{\partial t_{k-1}^2}|_{t^{\mathrm{flex}}} = 0$, such that: 
    \begin{enumerate}
        \item For $t_{k-1}<t^{\mathrm{flex}}$, the sensitivity of $\Phi$ to changes in $t_{k-1}$ should decrease with $t_{k-1}$ to prevent small variations in $t_{k-1}$ from causing large changes in $s_k$. The sensitivity of $\Phi$ to changes in $t_{k-1}$ should be lower the closer to $t^{\mathrm{min}}$.
        \item For $t_{k-1}>t^{\mathrm{flex}}$, the sensitivity of $\Phi$ should increase with $t_{k-1}$ to prevent small changes in $t_{k-1}$ from causing no change in $s_k$. The sensitivity of $\Phi$ should be higher the closer to $t^{\mathrm{max}}$.
    \end{enumerate}
\end{enumerate}

Additionally, we want $\Phi$ to operate satisfactorily with event processing algorithms with wide differences (several orders of magnitude) in $t_{k-1}$. Hence, we used $\log(t_{k-1})$ instead of $t_{k-1}$, where $\log$ represents the natural logarithm. The suggested $\Phi$ in the ASAP code is:
\begin{equation}
\Phi(t_{k-1}) = \arctan(\kappa \log(t_{k-1})).
\label{eq:my_fun}
\end{equation}

\begin{figure}[ht]
\begin{center}
\includegraphics[trim={0.2cm 0cm 1.8cm 0.4cm},clip,height=4cm,width=0.6\linewidth]{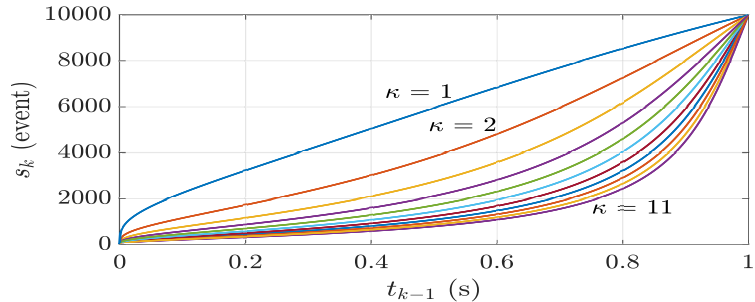}
\caption{Values of function $s_k = A\Phi(t_{k-1})+B$ for $\kappa \in [1,11]$ taking $t^{\mathrm{min}} = 10^{-6}$ s, $t^{\mathrm{max}} = 1$ s, $s^{\mathrm{min}} = 1$, and $s^{\mathrm{max}} = 10^4$. For these values, $A \approx 6.395 \cdot 10^3$ and $B = 10^4$.}
\label{fig:kappa}
\end{center}
\end{figure}

This function fulfills the above conditions, guarantees the existence of one inflection point, and includes only one parameter $\kappa$ that adjusts the curvature of $\Phi$ to set the desired inflection point. Given $\kappa$, the inflection point $t^{\mathrm{flex}}$ can be determined by a closed expression:
\begin{equation}
t^{\mathrm{flex}} = \exp\left(\frac{-\sqrt{\kappa^2 - 1}}{\kappa} - 1\right).
\label{eq:k}
\end{equation}
$|\kappa| < 1$ provides unfeasible solutions in Eq. (\ref{eq:k}). For $|\kappa| > 1$ there is a second inflection point $t^{\mathrm{flex}}_2$ in $\Phi$ when taking the two solutions of the  square root. It can be noticed that $t^{\mathrm{flex}}_2 > t^{\mathrm{flex}}$ and that $ \frac{\partial^2 \Phi(t_{k-1})}{\partial t_{k-1}^2} \approx 0$, $\forall t_{k-1} \in [t^{\mathrm{flex}}_2, t^{\mathrm{max}}]$. Hence, $t^{\mathrm{flex}}_2$ has negligible influence on the sensitivity of $\Phi$. The value of $\kappa$ modifies the shape of $\Phi$. Figure \ref{fig:kappa} shows the resulting curves $s_k = A\Phi(t_{k-1})+B$ when using different values of $\kappa \in [1,11]$ and computing  $A$ and $B$ with Eqs. (\ref{eq:a}) and (\ref{eq:b}) using $t^{\mathrm{min}} = 10^{-6}$ s, $t^{\mathrm{max}} = 1$ s, $s^{\mathrm{min}} = 1$, and $s^{\mathrm{max}} = 10^4$. These values are the roughly-selected upper and lower bounds for $t_{k-1}$ and $s_k$ that we observed in a wide variety of problems and event processing algorithms, and are proposed as default values in the ASAP code. Although they have been validated for all the problems tested, they can be fine-tuned if desired, see Section \ref{sec:results}.

The value of $\kappa$ determines the position of $t^{\mathrm{flex}}$ and hence, the point of operation of dynamic packaging. The default value for $\kappa$ in the ASAP code was roughly selected as $\kappa=5$, the center of the range shown in Fig. \ref{fig:kappa}. This value provided satisfactory results in all the problems that were tested, some of which are shown in Section \ref{sec:results}. Additionally, Section \ref{sec:results} briefly presents how to adapt $\kappa$ to specific conditions or applications.

To reduce the time to compute Eq. (\ref{eq:s_generic}), e.g., for interest in low-resource computers such as those used in Sections \ref{sec:expdrone} and \ref{sec:exppajaro}, $\Phi$ was approximated by its Taylor series expansion around an operating point $a$:
\begin{equation}
\Phi(t_{k-1}) \approx \sum_{j=0}^{N}{c_j (t_{k-1}-a)^j},
\label{eq:taylor}
\end{equation}
where $c_j$ is the $j$-th Taylor coefficient of $\Phi$ and $N$ is the order of the Taylor series expansion. Coefficients $c_j$ are tabulated for logarithmically-spaced values of $a$. For instance, using $N=3$ obtains a computational cost saving of $68.9\%$ w.r.t. obtaining $\Phi$ directly and an approximation error of $1.07 \cdot 10^{-6}$, whereas $N=5$ involves a computational cost saving of $60.41\%$ and an approximation error of $2.0247 \cdot 10^{-9}$.

\section{Gamma-Filter}
\label{sec:bucle1}
Adaptive packaging is insufficient for preventing algorithm processing overflow in case of high event rates. $\gamma$-filter reduces the number of events to be processed by randomly removing an adaptive percentage of the input events. It receives as input the event stream captured by the camera and outputs a filtered stream with a lower or equal number of events. Discarding events in cases with high event rates is not uncommon and has been proposed in several works. In our system, the event removal percentage is dynamically adapted depending on the current event rate and the value of $t_{k-1}$. Additionally, event removal should be performed efficiently to reduce the delay. We adopted random event removal. We performed preliminary experiments (Section \ref{sec:expbucle_gammafilter}) in which different types of event processing algorithms were fed with event streams resulting from filtering the input stream with different event removal percentages. We confirmed that with high event rates, the algorithm performance degradation was very low with event removal percentages of up to 80\%, and that removal was more effective (lower algorithm degradation and higher processing saving) with high event rates.

The operation of $\gamma$-filter depends on $\gamma_i \in \mathbb{R}: 0 \leq \gamma_i \leq 1$. For each received event $e_i$, a random number $\rho_i \in \mathbb{R} : 0 \leq \rho_i \leq 1$ is generated according to a uniform distribution $U(0,1)$. If $\rho_i < \gamma_i$, $e_i$ is removed. Otherwise, $e_i$ is provided as output. With $\gamma_i = 0$, all incoming events are removed; with $\gamma_i = 1$, all are given as output. $\gamma_i$ is selected proportionally to the incoming event rate. Let $e_i$ be an incoming event from the camera. Let $r_i \in \mathbb{R}^+$ be the current event rate when $e_i$ is received, measured by counting the events received during a sliding time window. The adaptation law makes $\gamma_i$ linearly dependent on $r_i$, which requires setting the maximum and minimum values of $r_i$, namely $r^{\mathrm{max}}_i$ and $r^{\mathrm{min}}_i$ respectively. Using static values for $r^{\mathrm{max}}_i$ and $r^{\mathrm{min}}_i$ would constrain adaptation to event rate changes. Hence, they are updated for each incoming event $e_i$ as follows:
\begin{eqnarray}
r^{\mathrm{max}}_i = \left\{\begin{array}{lcc}
\alpha r^{\mathrm{max}}_{i-1} & \mathrm{if} & r_i \leq r^{\mathrm{max}}_{i-1}\\
r_i & \mathrm{if} & r_i > r^{\mathrm{max}}_{i-1}
\end{array}\right.
,\quad 
r^{\mathrm{min}}_i = \left\{\begin{array}{lcc}
\dfrac{1}{\alpha}r^{\mathrm{min}}_{i-1} & \mathrm{if} & r_i \geq r^{\mathrm{min}}_{i-1}\\
r_i & \mathrm{if} & r_i < r^{\mathrm{min}}_{i-1}
\end{array}\right.
\label{eq:r_minmax_update}
\end{eqnarray}

If $r_i$ is lower than $r^{\mathrm{min}}_{i-1}$, the minimum value $r^{\mathrm{min}}_{i}$ is updated to $r_i$. Also, if $r_i$ is higher than $r^{\mathrm{max}}_{i-1}$, $r^{\mathrm{max}}_i$ is updated to $r_i$. Otherwise, $r^{\mathrm{min}}_i$ and $r^{\mathrm{max}}_i$ are computed from $r^{\mathrm{min}}_{i-1}$ and $r^{\mathrm{max}}_{i-1}$ using a forgetting factor $\alpha \in \mathbb{R} : 0 < \alpha \leq 1$. In addition, using a linear law to adapt $\gamma_i$ also requires establishing the upper and lower bounds for $\gamma_i$. The lower bound $\gamma^{\mathrm{min}}$ reflects the maximum event removal percentage, which, following the preliminary experiments, was taken as $\gamma^{\mathrm{min}}=0.2$. Using a fixed value for the minimum event removal percentage establishes a static relation between $r_i$ and $\gamma_i$. However, adapting it depending on the computational cost of the event algorithm is very interesting since it enables synergies between event packaging and $\gamma$-filter. The minimum event removal percentage $\hat{\gamma}_{k}$ is updated for each $t_{k-1}$ as follows:
\begin{equation}
\hat{\gamma}_k = \gamma^{\mathrm{max}} - \frac{t_{k-1} - t^{\mathrm{min}}}{t^{\mathrm{max}} - t^{\mathrm{min}}} (\gamma^{\mathrm{max}} - \gamma^{\mathrm{min}}),
\label{eq:gamma_max}
\end{equation}

where $t_{k-1}$, $t^{\mathrm{min}}$, and $t^{\mathrm{max}}$ are as defined in Section \ref{sec:bucle2}. $\gamma^{\mathrm{max}}$ is a fixed parameter that depends on the application, acts as the upper bound of $\hat{\gamma}_k$, and reflects the minimum event removal percentage. The value of $\hat{\gamma}_k$ depends on $t_{k-1}$ but always lies within $[\gamma^{\mathrm{min}}, \gamma^{\mathrm{max}}]$. 

The synergistic behaviour between both mechanisms can be clearly noticed. If the event algorithm becomes slower, $t_{k-1}$ will tend to $t^{\mathrm{min}}$. On one hand, adaptive packaging will increase the package size. On the other hand, according to Eq. (\ref{eq:gamma_max}), $\hat{\gamma}_k$ will tend to $\gamma^{\mathrm{min}}$, which will help select a lower value for $\gamma_k$ hence, avoiding the processing of a higher percentage of events, helping reduce the algorithm processing time. The opposite effect can be observed if $t_{k-1}$ tends to $t^{\mathrm{max}}$. Hence, using $t_{k-1}$ to adapt $\hat{\gamma}_k$ provides synergistic effects in which $\gamma$-filter helps event packaging in reducing the algorithm response time. 

Finally, the value of $\gamma_i$ is computed for each incoming event $e_i$ as follows:
\begin{equation}
\gamma_i = \hat{\gamma}_k - \frac{r_i - r^{\mathrm{min}}_i}{r^{\mathrm{max}}_i - r^{\mathrm{min}}_i} (\hat{\gamma}_k - \gamma^{\mathrm{min}}),
\label{eq:gamma}
\end{equation}
$\gamma_i$ adapts depending on $r_{i}$ and on $t_{k-1}$. This dual mechanism provides fast adaptation to changes in the event rate and in the algorithm response time.

We selected the random discard approach in $\gamma$-filter due to its extremely low computational burden. However, other filtering techniques can be used. For instance, spatio-temporal consistency filters could be used to discard events generated repeatedly in the same pixel or removing noisy events generated sporadically or in isolated regions on the image. Event filtering must be applied on each incoming event $e_i$, therefore, computationally complex filters may generate delays in cases with high event rates. Although the event random removal approach has been taken as default, different filters are provided in the ASAP code and the addition of new ones is straightforward.

\section{ASAP Adaptive Packaging Convergence}
\label{sec:conver}

Adaptive packaging selects a suitable package size $s_k$ for each processing temporal cost $t_{k-1}$. At the same time, the event algorithm processes the package of size $s_k$ in time $t_{k}$. It is necessary to demonstrate that adaptive packaging converges to feasible values of $s_k$ and $t_{k-1}$. Also, $\gamma$-filter does not disturb the relation between $s_k$ and $t_{k-1}$, since it only affects the number of events that are packaged, regardless of how they are packaged.

The convergence of adaptive packaging can be demonstrated as follows. First, assume that $t_{k} = g(s_{k})$, where $g: \mathbb{R}^+ \rightarrow \mathbb{R}^+$ models the time required by the algorithm to process the events in $p_k$. Function $g$ is assumed positive and strictly increasing in all its domain, i.e. $g(s_k) > 0$ and $\frac{\partial g(s_k)}{\partial s_k} > 0$, $\forall s_k > 0$. These two conditions are fulfilled in practice since the algorithm takes some time to process the package, and the higher number of events in the package, the longer it takes to process them. $t_k$ is used to select the size of the next package $s_{k+1}$. Let $s_{k+1} = f(t_{k})$, $f: \mathbb{R}^+ \rightarrow \mathbb{R}^+$ be the function used to select $s_{k}$. Using Eqs. (\ref{eq:s_generic}) and (\ref{eq:my_fun}), the $f$ function used in ASAP is:
\begin{equation}
s_{k+1} = f(t_{k}) = A \arctan{(\kappa \log(t_{k})) + B}.
\label{eq:f}
\end{equation}

Using Eqs. (\ref{eq:a}), (\ref{eq:b}), and (\ref{eq:k}), it can be deduced that $A \in \mathbb{R}^+$, $B \in \mathbb{R}$, and $\kappa \geq 1$. It can be also noticed that $f$ is strictly increasing $\forall t_k > 0$:
\begin{equation}
\frac{\partial f(t_k)}{\partial t_{k}} = \frac{A \kappa}{t_{k} (\kappa^2 \log^2(t_{k})+1)}.
\label{eq:dfdt}
\end{equation}

Assuming that $s_0>0$, it can be proved that the sequence of values of $s_k$ is either strictly increasing ($s_{k+1}>s_k$), strictly decreasing ($s_{k+1}<s_k$), or constant ($s_{k+1}=s_k$), $\forall k>0$. First, as $\frac{\partial g(s_{k})}{\partial s_{k}} > 0$, $\forall s_{k} > 0$, the larger the input $s_{k}$, the larger the output $t_{k}$. Second, from Eq. (\ref{eq:dfdt}), the larger the input $t_{k}$, the larger the output $s_{k+1}$. Consequently, if $a > b$, $f(g(a)) > f(g(b))$ and vice versa. Therefore:
\begin{enumerate}
    \item \textit{Constant}: Assuming $s_{k+1} = s_{k}$, the package size at $k+2$ is $s_{k+2} = f(g(s_{k+1})) = f(g(s_{k})) = s_{k+1}$. Hence, by induction, $s_{k+n} = s_{k}$, $\forall k > 0$, $\forall n > k$.    
    \item \textit{Strictly increasing}: Assuming $s_{k+1} > s_{k}$, the package size at $k+2$ is $s_{k+2} = f(g(s_{k+1})) > f(g(s_{k})) = s_{k+1}$. By induction, $s_{k+n} > s_{k}$, $\forall k > 0$, $\forall n > k$.
    \item \textit{Strictly decreasing}: Assuming $s_{k+1} < s_{k}$, the package size at $k+2$ is $s_{k+2} = f(g(s_{k+1})) < f(g(s_{k})) = s_{k+1}$. Hence, $s_{k+n} < s_{k}$, $\forall k > 0$, $\forall n > k$.
\end{enumerate}

\begin{figure}[ht]
\begin{center}
\includegraphics[trim={0cm 0cm 0cm 0cm},clip,height=4cm,width=0.6\linewidth]{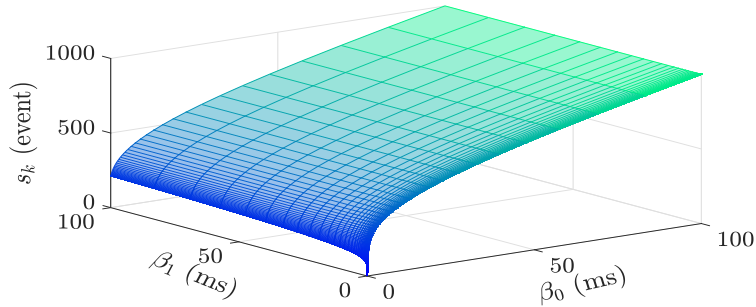}
\caption{Package size $s_k$ from numerical solution of $s_k = A \Phi(t_{k-1}) + B$ and $t_k = \beta_0 + \beta_1 s_k$ for $\beta_0, \beta_1 \in (0, 0.1]$ s using $s^{\mathrm{min}} = 1$, $t^{\mathrm{min}} = 10^{-6}$ s, $s^{\mathrm{max}} = 10^3$, and $t^{\mathrm{max}} = 0.1$ s.}
\label{fig:numerical_solution}
\end{center}
\end{figure}

Thus, it can be also proved that $s_k$ converges when $k \longrightarrow$ $\infty$ in any of the three cases. Case \textit{Constant}, all the values in the sequence are equal, so $s_k = s_0$ $\forall k$. Cases \textit{Strictly increasing} and \textit{Strictly decreasing}, as $f$ is bounded $-\frac{1}{2}A\pi + B < f(t_{k}) < \frac{1}{2}A\pi + B$, $\forall t_{k}>0$, $s_k$ has to converge, i.e., $\lim_{k \rightarrow \infty}{s_k} \in \mathbb{R}$. Otherwise, if $s_k$ diverged, then $|\lim_{k \rightarrow \infty}{s_k}| = \infty$. As an example, convergence is analyzed for $t_{k}=g(s_{k}) = \beta_0 + \beta_1 s_{k}$, where $\beta_0 \in \mathbb{R}^+$ is the time required by the algorithm to perform operations that do not depend on the number of events and $\beta_1 \in \mathbb{R}^+$ is the time required for processing each event in the package. Figure \ref{fig:numerical_solution} presents the values to which $s_k$ converges for $\beta_0, \beta_1 \in (0, 0.1]$ s using $s^{\mathrm{min}} = 1$, $t^{\mathrm{min}} = 10^{-6}$ s, $s^{\mathrm{max}} = 10^3$, and $t^{\mathrm{max}} = 0.1$ s.

\section{Experimental Results}
\label{sec:results}

ASAP is configured by only setting the bounds of $\gamma_i$, $t_{k-1}$, and $s_k$, and the value of $\kappa$. We suggest reasonable values (default in the ASAP code) that cover a wide variety of applications and event processing algorithms, see Table \ref{table:parameters}. These values exhibited the expected performance in the wide variety of tested applications and conditions, some of which are presented in this section.

\begin{table}[ht]
\begin{center}
\setlength\arrayrulewidth{1pt}
\begin{tabular}{|c|c|c|}
\hline
\cellcolor{gray!25}\textbf{Parameter} & \cellcolor{gray!25}\textbf{Description} & \cellcolor{gray!25}\textbf{Default value} \\ \hline
$\gamma^{\mathrm{min}}$ & Lower bound for $\gamma_k$ & $0.2$ \\ \hline
$\gamma^{\mathrm{max}}$ & Upper bound for $\gamma_k$ & $1$ \\ \hline
$t^{\mathrm{min}}$ & Lower bound for $t_{k-1}$ & $10^{-6}$ s \\ \hline
$t^{\mathrm{max}}$ & Upper bound for $t_{k-1}$ & $0.1$ s \\ \hline
$s^{\mathrm{min}}$ & Minimum package size & $1$ event \\ \hline
$s^{\mathrm{max}}$ & Maximum package size & $1000$ events \\ \hline
$\kappa$ & $\Phi$ factor see Eq. (\ref{eq:my_fun}) & $5$ \\ \hline
\end{tabular}
\end{center}
\caption{ASAP parameters and their default values.}
\label{table:parameters}
\end{table}

Manually setting these parameters enables fine-tuning ASAP for a specific problem. In cases with low event rates and efficient event processing algorithms that operate with packet sizes near to 1, it is interesting to select values of $s^{\mathrm{max}}$ and $t^{\mathrm{max}}$ lower than the default to provide ASAP with higher sensitivity when selecting $s_k$ with values close to 1. In these cases it is interesting to use higher values of $\kappa$ that set lower values of $t^{\mathrm{flex}}$, the point of operation of event packaging. Conversely, with high event rates and high time-consuming event algorithms it is interesting to set higher values of $s^{\mathrm{max}}$ and $t^{\mathrm{max}}$ than the default, and lower values of $\kappa$ to force ASAP to use larger packages as $t_{k-1}$ becomes larger.

ASAP can operate with any event processing algorithm. The only requirement is that the algorithm should feed back the time it required to process each received package. Such modification is minor and was easily performed in all the different algorithms we tested. ASAP was implemented in C++11 under ROS Melodic Morenia. It uses the \textit{libcaer} library for low-level event acquisition from the camera, but it can be integrated with other widely-used acquisition drivers (e.g. \textit{jAER}). 

First, both adaptive mechanisms are analyzed in Sections \ref{sec:expbucle_adaptivepackaging} and \ref{sec:expbucle_gammafilter}. In Sections \ref{sec:expdrone} and \ref{sec:exppajaro}, ASAP is validated and evaluated on board a quadrotor and a flapping-wing robot. In the analyses we use the \textit{delivery time} $\tau_k$ and \textit{building time} $\pi_k$ presented in Section \ref{sec:general}.

\begin{figure}[ht]
    \centering
    \includegraphics[height=3.5cm, width=0.3\linewidth]{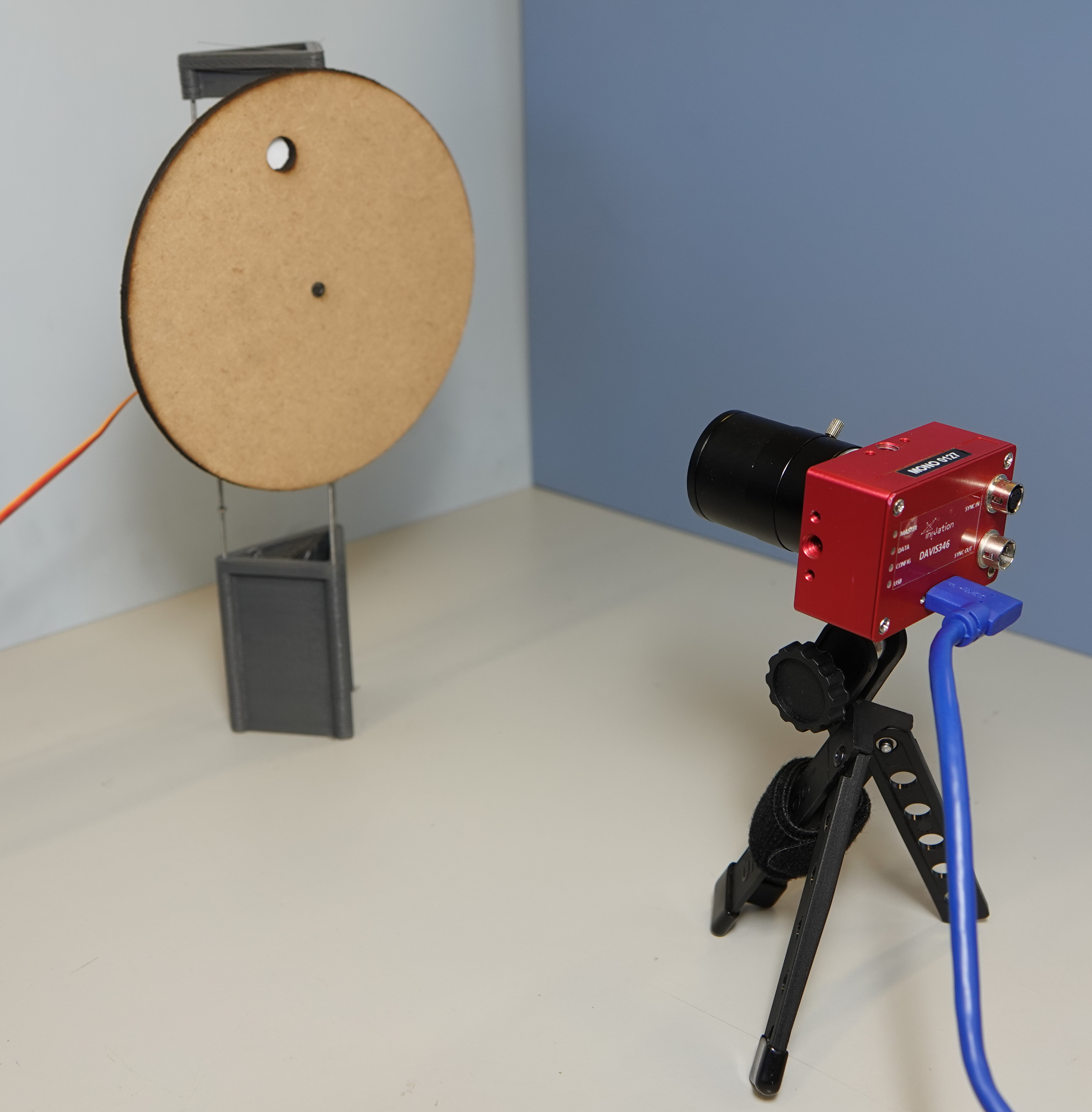}
    \includegraphics[height=3.5cm, width=0.3\linewidth]{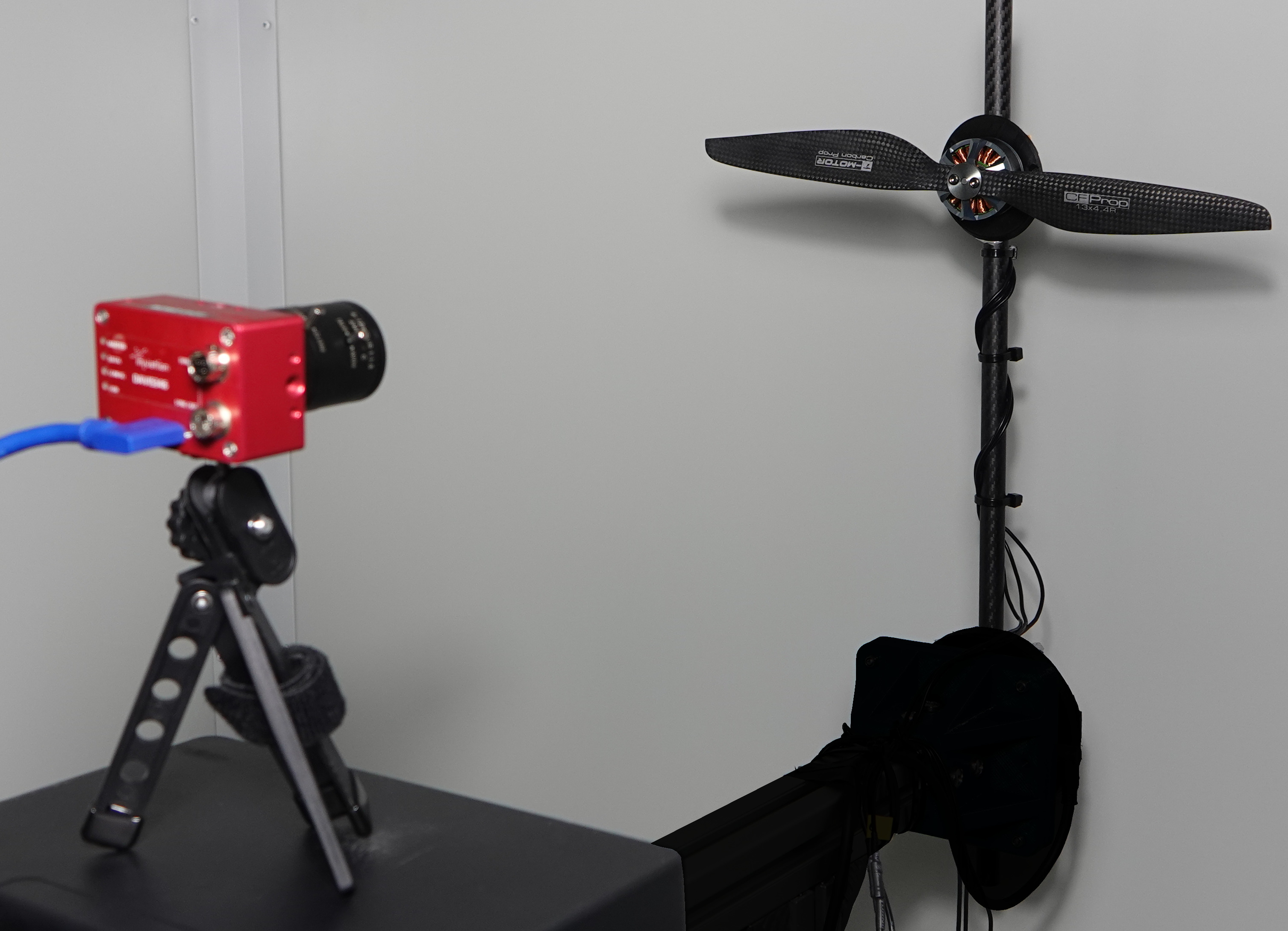}
    \includegraphics[height=3.5cm, width=0.6\linewidth]{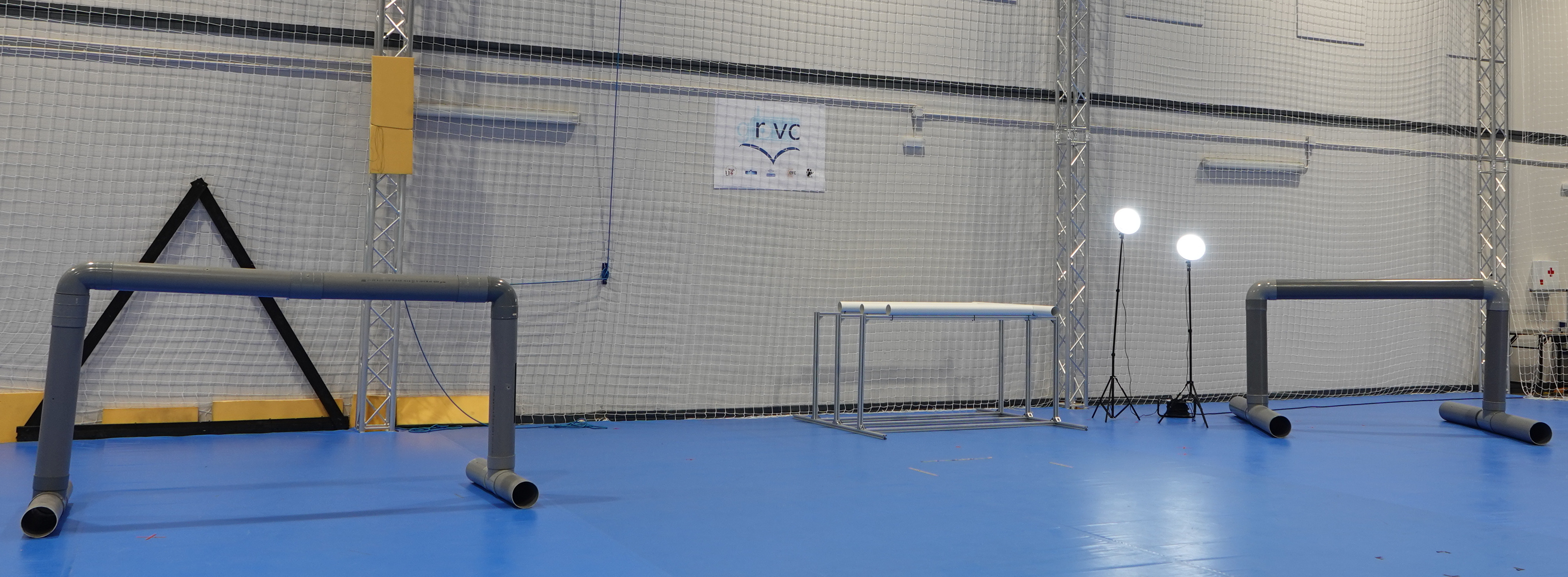}
    \caption{Experimental setups for validation of packaging (top-left), $\gamma$-filter (top-right), and full ASAP scheme (bottom).}
    \label{fig:experiments_setup}
\end{figure}

\subsection{Evaluation of Adaptive Packaging}
\label{sec:expbucle_adaptivepackaging}

First, we analyze the isolated effect of adaptive packaging. A DAVIS346 event camera was placed in front of a spinning wheel (see Fig. \ref{fig:experiments_setup}-top-left) that rotated at a constant angular velocity providing an approximately constant event rate of $\sim 70$ events/ms.
 
\begin{figure}[ht]
    \centering
    \includegraphics[trim={0.3cm 1cm 2cm 0.9cm},clip,height=6cm, width=0.6\linewidth]{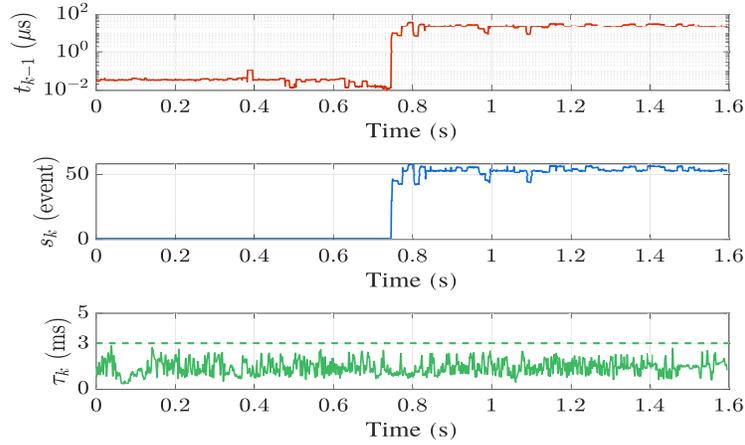}
    \caption{Values of $t_{k-1}$ (a), $s_k$ (b), and $\tau_k$ (c) resulting in an experiment in which the algorithm computational cost changed at $\sim 0.75$ s from $O(1)$ to $O(25)$.}
    \label{fig:exp_complexity_step}
\end{figure}

\begin{figure}[ht]
\begin{center}
\includegraphics[trim={0.6cm 0.1cm 1.4cm 0.3cm},clip,height=4cm,width=0.6\linewidth]{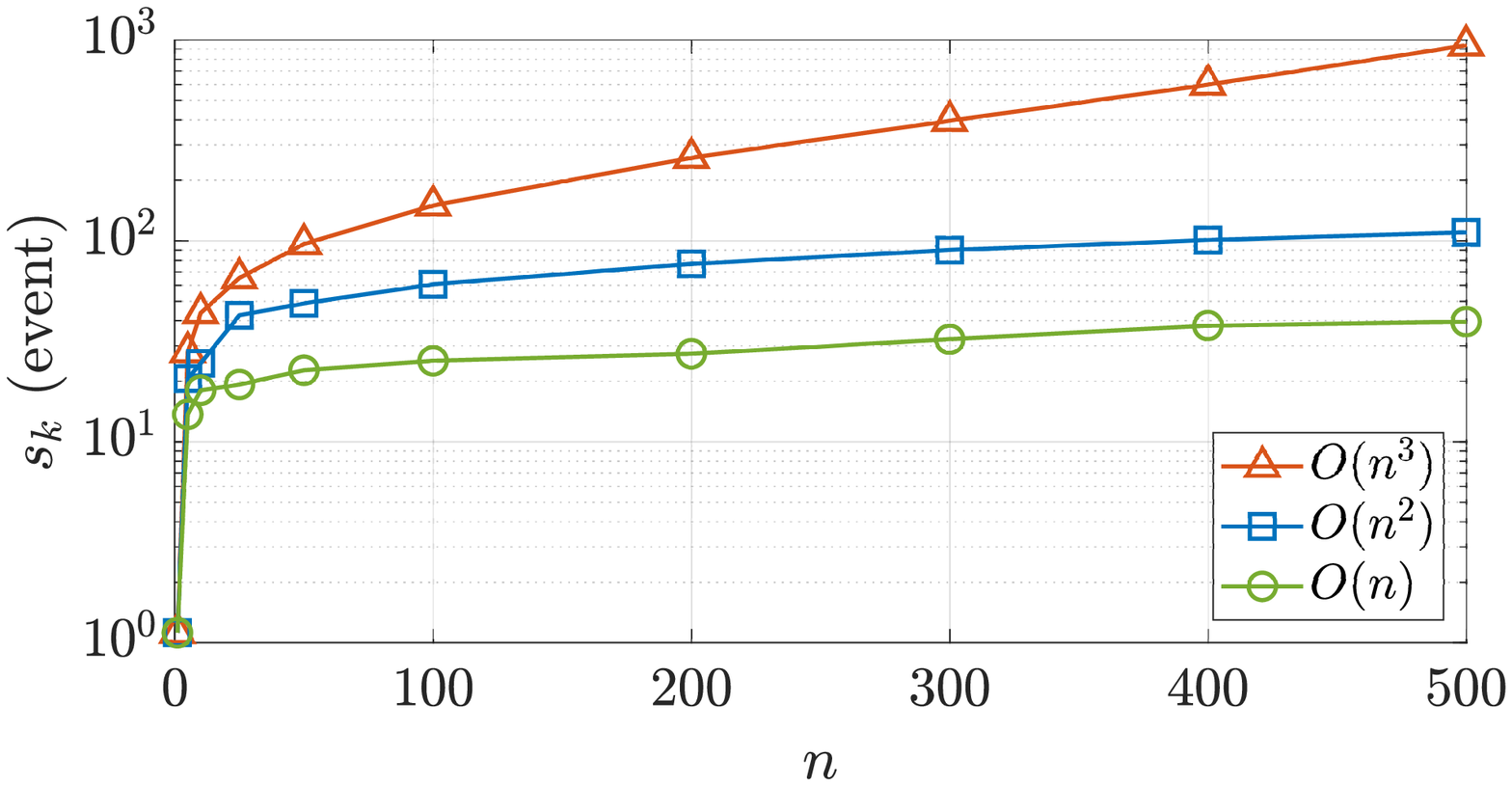}
\caption{Mean package sizes $s_k$ for event processing algorithms with different computational complexity.}
\label{fig:exp_complexity_simple}
\end{center}
\end{figure}

Figure \ref{fig:exp_complexity_step} shows its behaviour when there are sudden changes in $t_{k-1}$. Before $\sim 0.75$ s an efficient event processing algorithm with computational cost $O(1)$ was used, and adaptive packaging selected $s_k \approx 1$. At $\sim 0.75$ s, the computational cost of the algorithm was increased to $O(25)$. It can be seen in Fig. \ref{fig:exp_complexity_step}-b that $s_k$ adapted with no overshoot and rapidly reached the steady state. We define $\nu$ as the number of iterations required by $t_{k-1}$ to reach an steady-state error $< \pm 1\%$ w.r.t. the mean steady-state value. In this experiment, ASAP adapted in only $\nu = 5$ iterations. The resulting package \textit{delivery times} $\tau_k$ shown in Fig. \ref{fig:exp_complexity_step}-c were always lower than 3 ms and had similar values in both algorithms, meaning that there were not computational overflows despite the high differences in $t_{k-1}$ between both algorithms. In case of computational overflow, the packages would be buffered and $\tau_k$ would increase over time. Algorithms with different values of $t_{k-1}$ will result in different $s_k$. 
Fig. \ref{fig:exp_complexity_simple} shows the mean values of $s_k$ when using event algorithms with linear, quadratic, and cubic complexity.

\begin{figure}[ht]
    \centering
    \includegraphics[trim={4cm 1cm 4.2cm 1.2cm},clip,height=8cm,width=0.95\linewidth]{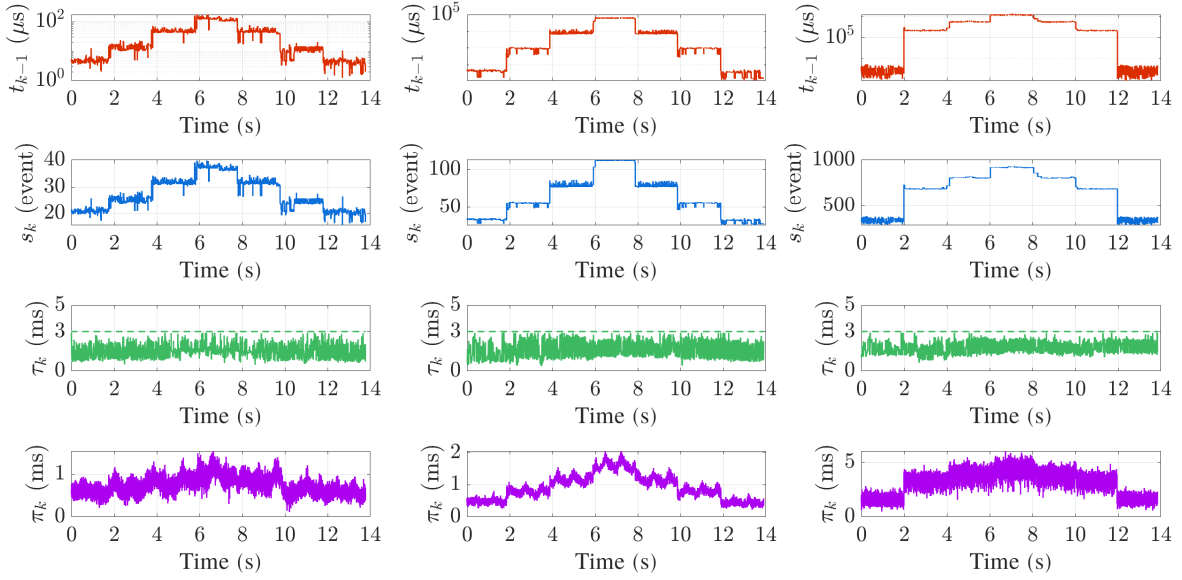}
    \caption{Values of: $t_{k-1}$ (a), $s_k$ (b), $\tau_k$ (c), and $\pi_k$ (d) when changing the algorithm computational cost with linear (left), quadratic (center), and cubic (right) complexity. Every $\sim 2$ s $n$ changed following the sequence $10$, $50$, $200$, $500$, $200$, $50$, $10$.}
    \label{fig:exp_complexity_3}
\end{figure}

Figure \ref{fig:exp_complexity_3} shows the performance of adaptive packaging to changes in the computational cost of event algorithms with linear $O(n)$ (left), quadratic $O(n^2)$ (center), and cubic $O(n^3)$ (right) complexity. In each case, $n$ is changed every $\sim 2$ s following the sequence $10$, $50$, $200$, $500$, $200$, $50$, and $10$. The figure shows the evolution in the values of $t_{k-1}$ (Fig. \ref{fig:exp_complexity_3}-a), $s_k$ (Fig. \ref{fig:exp_complexity_3}-b), $\tau_k$ (Fig. \ref{fig:exp_complexity_3}-c), and $\pi_k$ (Fig. \ref{fig:exp_complexity_3}-d). Adaptive packaging successfully adapted to every change in a low number of iterations $\nu$, shown in Table \ref{table:exp_complexity_3}. Note that the more computationally expensive the algorithm, the higher the value of $s_k$ and hence $\pi_k$. In all cases, $\tau_k$ remained lower than 3 ms despite the high difference in $s_k$, meaning that there were no increasing waiting buffers and packages were processed as soon as they arrived. The values of $\lambda_k$ are not shown. As $\lambda_k = \tau_k + \pi_k$, \textit{latency} can be obtained from Figs. \ref{fig:exp_complexity_3}-c and \ref{fig:exp_complexity_3}-d. The values of $\tau_k$ had high similarity for all packages, hence, the main changes in $\lambda_k$ were caused by the different values of $\pi_k$. This behaviour means that ASAP increases \textit{latency} only in cases where the algorithm requires it, and not due to increasing waiting buffers.

The value of $\pi_k$ depends on the performance of ASAP ($s_k$) and the environment ($r_i$). The influence of ASAP in $\pi_k$ can be easily noticed in this experiment since the values of $r_i$ are approximately constant along this experiment. This influence is not that clear in the experiments below since $r_i$ has wide variations, and for clarity $\pi_k$ is not shown. The objective of ASAP is to keep $\tau_k$ steady regardless of the environment and the algorithm processing time. If $\tau_k$ remains approximately constant, it means that the packages were processed as soon as they arrive. Hence, $\tau_k$ is the most critical metric in the performance of our method and its analysis is the main focus in the experiments below.

\begin{table}[ht]
\begin{center}
\setlength\arrayrulewidth{1pt}
\begin{tabular}{|c|c|c|c|}
\hline
\cellcolor{gray!25}\diagbox[linewidth=0.65pt,width=6em]{\textbf{n}}{\textbf{O}} & \cellcolor{gray!25}\textbf{O(n)} & \cellcolor{gray!25}\textbf{O(n$^2$)} & \cellcolor{gray!25}\textbf{O(n$^3$)} \\ \hline
\cellcolor{gray!25}\textbf{10 $\mathbf{\rightarrow}$ 50} & 5 & 7 & 4 \\ \hline
\cellcolor{gray!25}\textbf{50 $\mathbf{\rightarrow}$ 200} & 6 & 3 & 4 \\ \hline
\cellcolor{gray!25}\textbf{200 $\mathbf{\rightarrow}$ 500} & 8 & 4 & 2 \\ \hline
\cellcolor{gray!25}\textbf{500 $\mathbf{\rightarrow}$ 200} & 5 & 3 & 5 \\ \hline
\cellcolor{gray!25}\textbf{200 $\mathbf{\rightarrow}$ 50} & 4 & 2 & 4 \\ \hline
\cellcolor{gray!25}\textbf{50 $\mathbf{\rightarrow}$ 10} & 3 & 2 & 3 \\ \hline
\end{tabular}
\end{center}
\caption{Number of iterations $\nu$ required by $t_{k-1}$ to reach an steady-state error $< \pm 1\%$ in the tests shown in Fig. \ref{fig:exp_complexity_3}.}
\label{table:exp_complexity_3}
\end{table}

\subsection{Evaluation of Gamma-Filter}
\label{sec:expbucle_gammafilter}

Next, the isolated effect of $\gamma$-filter is analyzed. The DAVIS346 was placed in front of a UAV propeller attached to a brush-less motor (see Fig. \ref{fig:experiments_setup}-top-right). The motor angular velocity was varied between 450 rpm (providing $\sim 500$ event/ms in average) and 4300 rpm ($\sim 8000$ event/ms). First, we analyze its performance with changes only in the input event rate $r_i$, and then, changes simultaneously in $r_i$ and $t_{k-1}$.

\begin{figure}[ht]
\begin{center}
\includegraphics[trim={0.1cm 0.5cm 1.9cm 0cm},clip,height=4.8cm,width=0.6\linewidth]{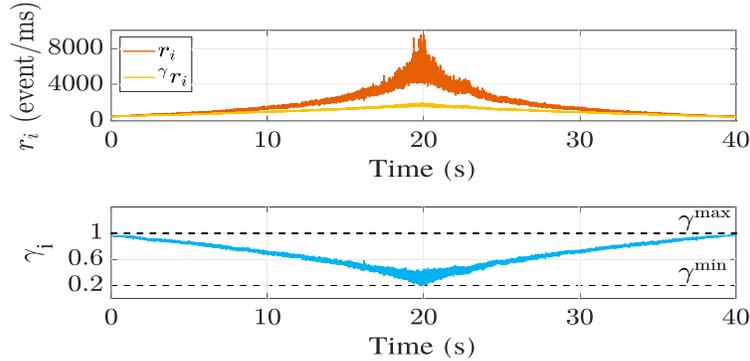}
\caption{Operation of $\gamma$-filter when the motor accelerated until $\sim 20$ s, and decelerates until $\sim 40$ s: top) input and output event rates from $\gamma$-filter; bottom) resulting values of $\gamma_i$.}
\label{fig:exp_gamma_1}
\end{center}
\end{figure}

Figure \ref{fig:exp_gamma_1} shows the operation of $\gamma$-filter when the motor accelerated from 450 to 4300 rpm and then decelerated to 450 rpm. The value of $t_{k-1}$ was kept constant. $\gamma$-filter dynamically adapted $\gamma_i$ to $r_i$, see Fig. \ref{fig:exp_gamma_1}-bottom. Figure \ref{fig:exp_gamma_1}-top shows the event rate in orange and the resulting event rate output from $\gamma$-filter in yellow.

\begin{figure}[ht]
\begin{center}
\includegraphics[trim={0.1cm 0.5cm 1.9cm 0.9cm},clip,height=6.8cm,width=0.6\linewidth]{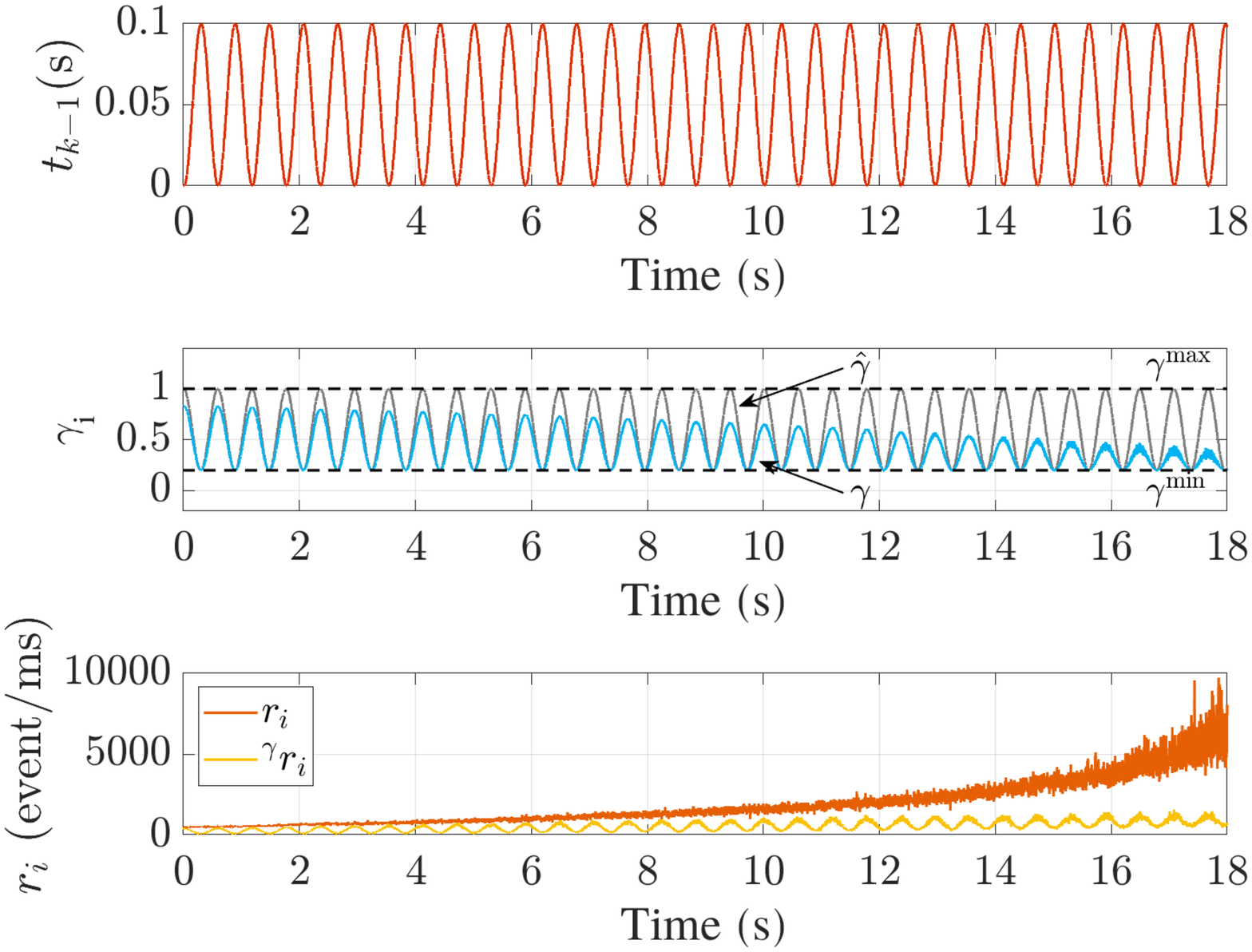}
\caption{Performance of $\gamma$-filter when the motor velocity increased from 450 to 4300 rpm, and simultaneously $t_{k-1}$ had sinusoidal oscillations between $t^{\mathrm{min}}=10^{-6}$ s and $t^{\mathrm{max}}=0.1$ s: top) $t_{k-1}$; center) resulting values for $\hat{\gamma}_k$ and $\gamma_i$; bottom) input and output event rates showing the filtering effect.}
\label{fig:exp_gamma_2}
\end{center}
\end{figure}

Figure \ref{fig:exp_gamma_2} shows an experiment in which the motor angular velocity was increased from 450 rpm to 4300 rpm and, simultaneously, $t_{k-1}$ had sinusoidal oscillations between $10^{-6}$ and $0.1$ s, see Fig. \ref{fig:exp_gamma_2}-top. The values of $\hat{\gamma}_k$ adapted to $t_{k-1}$ through Eq. (\ref{eq:gamma_max}), and the oscillations in $t_{k-1}$ induced oscillations between $\gamma^{\mathrm{min}}=0.2$ and $\gamma^{\mathrm{max}}=1$, see Fig. \ref{fig:exp_gamma_2}-center. Also, using Eq. (\ref{eq:gamma}), $\gamma_i$ decreases the higher the value of $r_i$ and varied between $\gamma^{\mathrm{min}}$ and $\hat{\gamma}_k$ depending on $r_i$. This effect can be noticed in Fig. \ref{fig:exp_gamma_2}-bottom. A very low percentage of events are removed with low values of $r_i$ and $t_{k-1}$, e.g. at the start of the experiment. A higher percentage of events are removed with higher values of $t_{k-1}$ and $r_i$, e.g. at $18$ s. $\gamma$-filter only removed a moderate number of events when event rates compromised the algorithm processing capacity.

\subsection{Evaluation on Board a Quadrotor Robot}
\label{sec:expdrone}

This section analyzes the operation of ASAP on board a quadrotor robot while navigating at different velocities. Figure \ref{fig:robots} shows the experimental platform: a \textit{DJI Flamewheel F450} frame equipped with a front-facing DAVIS346 event camera and a \textit{PixRacer} autopilot. Event onboard computation was performed using a low-cost Khadas VIM3 board, which included an Amlogic A311D SoC, $2$GB RAM, and $16$GB eMMC, mounting Ubuntu 18. The scenario was the GRVC Robotics Lab indoor flight arena endowed with a motion capture system with $24$ \textit{OptiTrack Prime$^x 13$} cameras that provided millimeter-accuracy pose estimations.

\begin{figure}[ht]
\centering
\includegraphics[trim={1.7cm 1cm 2cm 0.5cm},clip,height=8.5cm,width=0.64\linewidth]{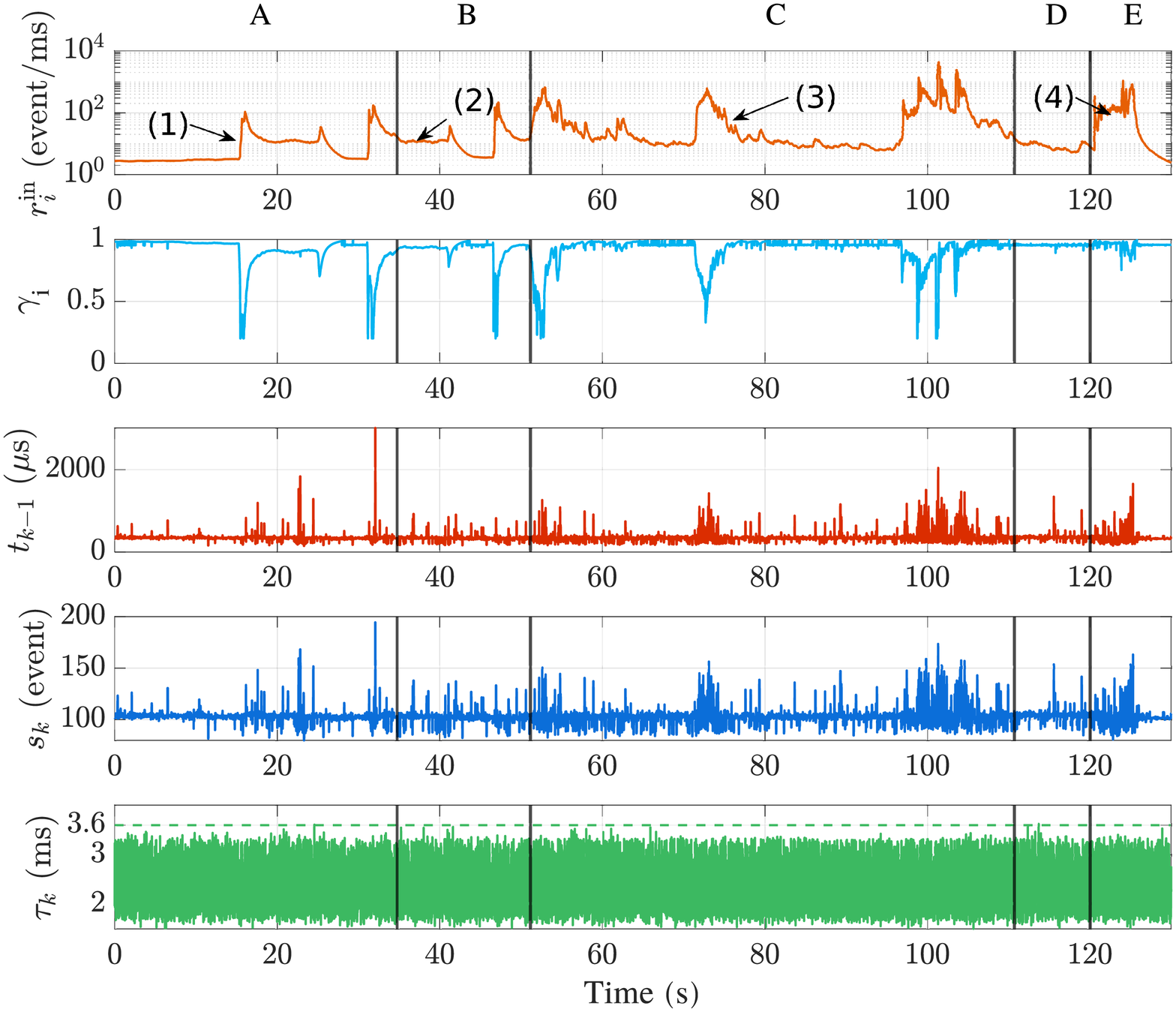}
\includegraphics[trim={3.5cm 2cm 3cm 0.5cm},clip,height=8.5cm,width=0.34\linewidth]{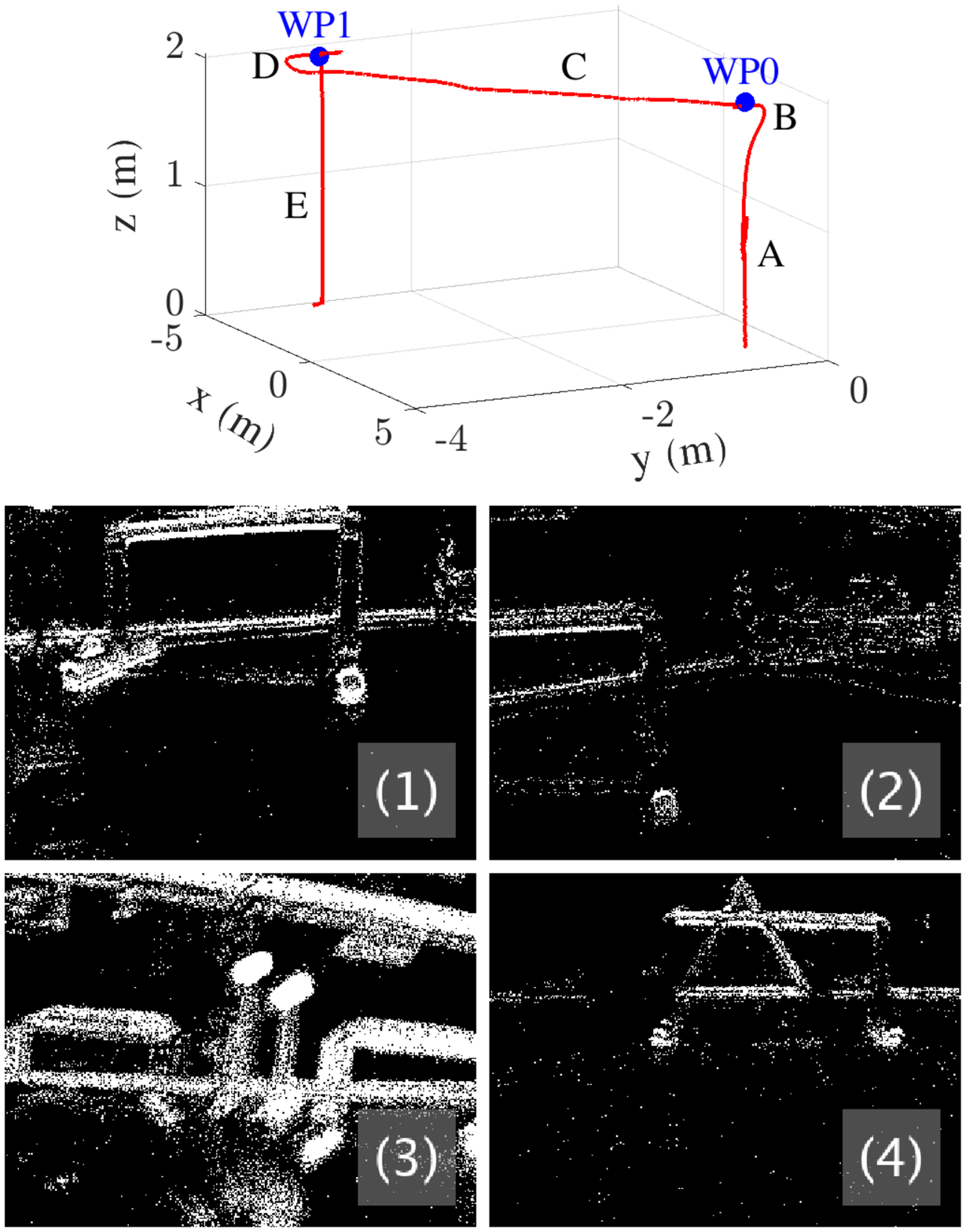}
\caption{Quadrotor experimental results. Left, from top to bottom: Values of $r_i$, $\gamma_i$, $t_{k-1}$, $s_k$, and $\tau_k$. Right-top: Quadrotor robot trajectory. Right-bottom: \textit{event images} (grouping events during $30$ ms) at times (1), (2), (3), and (4).}
\label{fig:exp_uav}
\end{figure}

In this experiment, ASAP fed the eHarris$^\ast$ from \citep{vasco2016fast} using the implementation from \citep{alzugaray2018asynchronous}. We use a corner detector for validation since it is dependent on the events generated in the neighborhood of each corner. This algorithm is suitable to properly evaluate the influence of $\gamma$-filter since its performance is particularly sensitive to event removal. The robot flight was designed to combine aggressive and slow movements. Each flight had $5$ stages: A) take-off; B) hovering at $WP0$ defined at 2 m over the take-off position; C) navigation to waypoint $WP1$ combining aggressive and smooth maneuvers; D) hovering at $WP1$; and E) landing. Figure \ref{fig:exp_uav}-left shows the event rates (in logarithmic axis) along the experiment, which suffered wide variations including: low event rates during stages B and D; average event rates in stages A and E; and sudden event rate peaks in stage C. Figure \ref{fig:exp_uav}-bottom-right shows \textit{event images} (grouping the events during $30$ ms) extracted from stages A, B, C, and E. Figure \ref{fig:exp_uav}-left shows (from top to bottom) the input event rates, $\gamma_i$, $t_{k-1}$, $s_k$, and $\tau_k$ along one experiment.

At stage A the quadrotor took off at $\sim 16$ s. The input event rates changed from very low values (before take-off) to moderate values with sudden peaks of up to $\sim 250$ event/ms. $\gamma$-filter reacted to these peaks by decreasing $\gamma_i$. Simultaneously, an increase in $r_i$ involved a higher number of events to be processed, and to avoid an increase in the delay, adaptive packaging increased $s_k$. As a consequence, $t_k$ increased. Both effects originated package delivery times of approximately constant values and always lower than 3.6 ms, regardless of the changes in $r_i$. At stage B the quadrotor stayed at stationary flight at $WP0$ but the scene contained moving objects that triggered events causing event rate changes. To adapt to these changes, $\gamma$-filter decreased $\gamma_i$ and $s_k$ increased to compensate delay increments. 

At stage C the quadrotor flew from $WP0$ to $WP1$ at a constant velocity except for three local accelerations (and decelerations) at times $53$ ($54$), $73$ ($75$), and $97$ ($104$) s. During accelerations, the input event rate changes were compensated by decreasing $\gamma_i$ and increasing $s_k$. Recall from Section \ref{sec:bucle1} that $\gamma$-filter keeps track of $r^{\mathrm{min}}$ and $r^{\mathrm{max}}$, which influence the computation of $\gamma_i$ in Eq. (\ref{eq:gamma}). The aggressive accelerations at time $97$ s, caused a strong increase in $r^{\mathrm{max}}$. From that moment on, $\gamma$-filter adapted to event rate changes with lower decrements in $\gamma_i$, which can be noticed at time $124$ s. The responsability to compensate event rate changes is shared between $\gamma$-filter and adaptive packaging. As it acquires more information on the problem (e.g. updating $r^{\mathrm{min}}$ and $r^{\mathrm{max}}$), it better adjusts the share between both mechanisms. During decelerations, the event rate reductions were compensated by increasing $\gamma_i$ and decreasing $s_k$.

At stage D the quadrotor stayed in stationary flight and there were no moving objects in the scene. There were low changes in $r_i$, and $\gamma_i$ was kept constant. Changes in $s_k$ were only caused by changes in the algorithm processing time. At stage E the quadrotor landed at $\sim 124$ s. The event rate had abrupt changes, caused by the robot contact with the ground. $\gamma$-filter attenuated this event rate peak by decreasing $\gamma_i$.

\begin{figure}[ht]
\begin{center}
\includegraphics[trim={2.2cm 1cm 4cm 0.2cm},clip,height=5cm,width=0.6\linewidth]{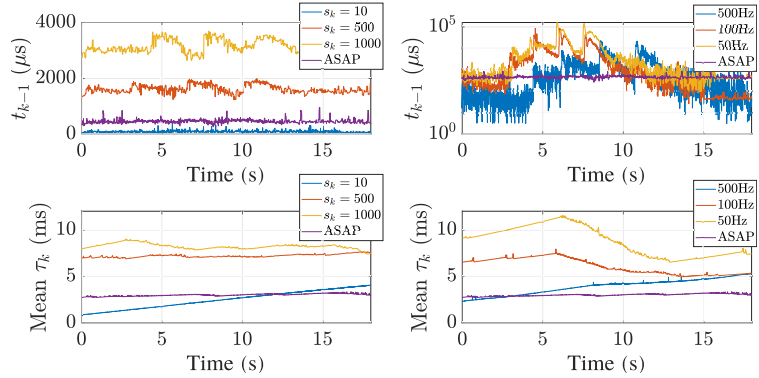}
\caption{Comparison of ASAP versus other event packaging methods: left) fixed size packages with $10$, $500$, and $1000$ events, and right) fixed time windows with $50$, $100$, and $500$ Hz: top) $t_{k-1}$, and bottom) $\tau_k$.}
\label{fig:exp_comparative_uav}
\end{center}
\end{figure}

Despite the wide variations in the event rate along the experiment, ASAP successfully adapted $\gamma_i$ and $s_k$ so that $\tau_k$ remained approximately constant and lower than $3.6$ ms, meaning that the algorithm did not suffer overflows. The suggested ASAP parameters performed as expected, also during stage C where the robot accelerations caused event rate peaks of $5000$ event/ms. Figure \ref{fig:exp_comparative_uav} compares the performance of ASAP versus static event packaging methods while flying from $WP0$ to $WP1$: left) fixed size packages with $10$, $500$, and $1000$ events per package, and right) fixed time window packages at $50$, $100$, and $500$ Hz. ASAP provided values of $t_{k-1}$ significantly lower than using fixed size packages of $500$ and $1000$ events. Only fixed-size packages of $10$ events were processed faster. However, in this case, the value of $\tau_k$ increased along the experiment, which means that the packages were queued in increasing waiting buffers and the algorithm suffered overflows. Similarly, ASAP provided lower $t_k$ than packaging with fixed time windows at $50$ and $100$ Hz, and package delivery times increased over time when packaging at $500$ Hz, which involved algorithm overflows.

Each event algorithm, depending on the type and cost of the involved processing, has a range of package sizes and frequencies that keep the algorithm responsiveness preventing overflows. Manually selecting the value of the static package size/frequency can involve long trial-and-error iterative processes, and the found value might be unsuitable if the event rate changes. The efficacy of the event algorithm fed with ASAP was also evaluated and compared. Three areas in the scenario were selected in the evaluation: the upper corner of the triangle in Fig. \ref{fig:experiments_setup}-bottom (\textit{Triangle}), a spotlight (\textit{Spotlight}), and the end of a pipe (\textit{Pipe}). Performance was estimated by the \textit{corner detection rate}, defined as the ratio between the time at least one corner was detected at these areas and the total time the areas were within the camera's field of view. Table \ref{table:eharris_metrics} presents the average \textit{corner detection rate} obtained for the different event packaging methods w.r.t. the best rate. In all the cases feeding with ASAP provided results $>95\%$. A more thorough analysis could be performed to in-detail evaluate the influence of $\gamma$-filter in different evaluation metrics of Harris detector. However, that is not the objective of this paper and we preferred to use a general performance metric for brevity. These results validate that ASAP or its random event removal $\gamma$-filter do not perturb the efficacy of the event-based corner detector tested. ASAP has also been validated with other \textit{event-by-event} algorithms with different computational burden, such as \citep{mueggler2017fast}, and other algorithms based on \textit{event images} and \textit{hybrid} processing, such as \citep{gomez2020asynchronous}, \citep{martinez2020towards}, or \citep{rodriguez2021uav}.

\newcommand{\z}{\phantom{0}}
\begin{table}[ht]
\begin{center}
\setlength\arrayrulewidth{1pt}
\renewcommand\arraystretch{1.25}
\setlength{\tabcolsep}{0.3em}
\arrayrulecolor{black}
\begin{tabular}{|c|c|c|c|c|}
\hhline{~~|---|}
\multicolumn{2}{c|}{} & \cellcolor{gray!25}\textbf{\textit{Triangle}} & \cellcolor{gray!25}\textbf{\textit{Spotlight}} & \cellcolor{gray!25}\textbf{\textit{Pipe}} \\ \hline
\cellcolor{gray!25}&\cellcolor{gray!25}\textbf{10 events}   & \z86.16\% & \z84.24\% & \z85.33\% \\ \hhline{|>{\arrayrulecolor{gray!25}}->{\arrayrulecolor{black}}|----|}
\cellcolor{gray!25}&\cellcolor{gray!25}\textbf{500 events}  & \z91.20\% & \z96.82\% & \z91.47\% \\ \hhline{|>{\arrayrulecolor{gray!25}}->{\arrayrulecolor{black}}|----|}
\multirow{-3}{*}{\cellcolor{gray!25}\shortstack[l]{\textbf{Fixed-size}\\\textbf{packaging}}}&\cellcolor{gray!25}\textbf{1000 events} & \z95.66\% & \z98.73\% & \z96.70\% \\ \hline
\cellcolor{gray!25}&\cellcolor{gray!25}\textbf{50 Hz}                 & \z95.59\% & \z95.32\% & \z94.30\% \\
\hhline{|>{\arrayrulecolor{gray!25}}->{\arrayrulecolor{black}}|----|}
\cellcolor{gray!25}&\cellcolor{gray!25}\textbf{100 Hz}                &  100.00\% &  100.00\% &  100.00\% \\
\hhline{|>{\arrayrulecolor{gray!25}}->{\arrayrulecolor{black}}|----|}
\multirow{-3}{*}{\cellcolor{gray!25}\shortstack[l]{\textbf{Fixed-time}\\\textbf{packaging}}}&\cellcolor{gray!25}\textbf{500 Hz}                & \z93.89\% & \z93.91\% & \z93.26\% \\ \hline
\multicolumn{2}{|      c|}{\cellcolor{gray!25}\textbf{ASAP}}                  & \z95.10\% & \z97.47\% & \z95.13\% \\ \hline
\end{tabular}
\end{center}
\caption{Performance degradation w.r.t. the best corner detection efficacy when comparing ASAP with static packaging.}
\label{table:eharris_metrics}
\end{table}

The influence of $\gamma$-filter on the algorithm performance depends on the type of event processing. Event-based algorithms relying on local features --e.g., corners-- will be particularly sensitive to random event removal since the result of each pixel depends on the consistency in the event generation of its neighborhood. Conversely, algorithms based on the use of non-local information --e.g., lines or blob detection-- will be more robust to random discard. Random event removal has been validated with different types of algorithms, such as clustering \citep{tapia2020asap}, line detection \citep{gomez2021why}, and optical flow estimation \citep{rodriguez2022free}, where it operated satisfactorily involving no significant performance loss. The results are not shown for brevity.

\subsection{Validation on Board a Flapping-Wing Robot}
\label{sec:exppajaro}

ASAP is particularly interesting for onboard perception for flapping-wing robots. First, ornithopters suffer from mechanical vibrations and wide abrupt movements due to the flapping strokes, which originate large event rates \citep{rodriguez2021griffin}. They also have gliding capabilities and their flights often combine flapping and gliding modes, involving wide event rate changes. Besides, they have strict payload and energy limitations, which severely constrain the onboard processing hardware. Furthermore, dealing with their high vibration level can require complex event processing algorithms. ASAP was validated onboard the \textit{E-Flap} ornithopter robot developed at the GRVC Robotics Lab \citep{zufferey2021design}. \textit{E-Flap} has an empty weight of 510 g, a maximum payload of 520 g, a total length of 95 cm and a maximal wing span of 1.5 m. It was equipped with a DAVIS346 event camera and a Khadas VIM3 for onboard perception mounting Ubuntu 18.

\begin{figure}[ht]
\begin{center}
\includegraphics[trim={0.1cm 1.2cm 1.8cm 0.05cm},clip,height=6.5cm,width=0.6\linewidth]{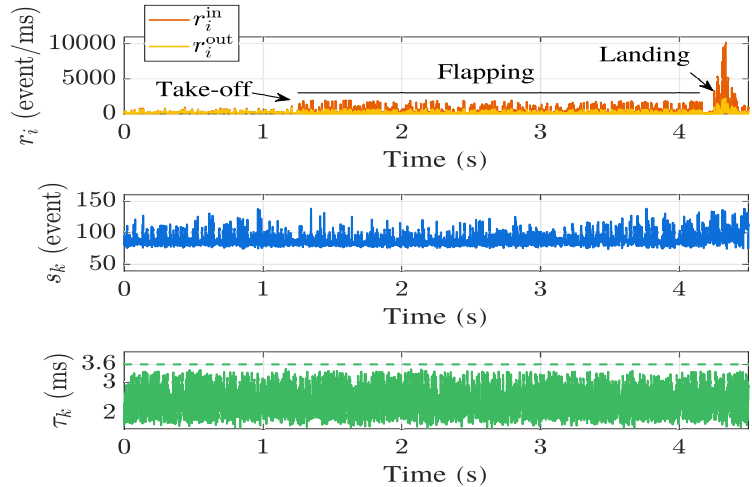}
\caption{\textit{E-Flap} experimental results. From top to bottom: Values of $r_i$, $s_k$, and $\tau_k$.}
\label{fig:exp_ornithopter}
\end{center}
\end{figure}

Figure \ref{fig:exp_ornithopter} shows the results obtained when using ASAP to feed an event-based guidance algorithm. The algorithm detected lines combining \textit{event image} based line detection with \textit{event-by-event} line tracking. The line estimates, obtained at rates $>$100 Hz, were input to a closed-loop visual servoing scheme that controlled the ornithopter tail deflectors to guide \textit{E-Flap} to a pre-selected final position. Vibrations caused by flapping strokes, aggressive maneuvers, and landing contact caused abrupt event rate changes, which were attenuated by $\gamma$-filter, as can be seen in Fig. \ref{fig:exp_ornithopter}-top. Due to adaptive packaging, the responsiveness of the algorithm (which was executed on board in a low-resource Khadas VIM3) was not affected by the event rate changes and did not suffer computational overflows, see Fig. \ref{fig:exp_ornithopter}-center. As can be noticed in Fig. \ref{fig:exp_ornithopter}-bottom, the package delivery times were similar along the flight and always lower than 3.6 ms. ASAP fed the events to the event-based guidance method, which consistently reached the target position with an error $<$0.4 m. The above results validate the use of ASAP in this challenging problem.

\section{Conclusions and Future Work}
\label{sec:conclusions}

Feeding event processing algorithms with static event packaging (either with constant size or rate) has limitations when dealing with the strong changes in the event rate and in the processing algorithm efficiency that can be found in robotics applications. This paper presented ASAP, a framework for event handling that dynamically adapts event delivery --absorbing changes in the event rates and in the algorithm event processing times-- to enhance processing responsiveness and prevent overflow. It is composed of two adaptive feedback-based mechanisms. The first one prevents event processing overflows by randomly discarding an adaptive percentage of the incoming events. The second mechanism dynamically adapts the size of the \textit{event packages}. ASAP has guaranteed convergence and is flexible to the processing algorithm and its computational complexity. It has been validated in different experiments and scenarios, including experiments with one quadrotor and one ornithopter robot, where ASAP delivers events to different online event algorithms that were executed on board in low-resource embedded computers.

ASAP paves the way for the use of online event vision techniques in a wide range of applications. Although the suggested $\Phi$ function and parameters have shown their efficacy in all the experiments performed, using other functions and parameters is still an open field for research. The extension of ASAP to a full-featured framework for event-based vision research, including camera calibration, event stream analysis, and visualization tools is object of current development.

\section*{Acknowledgment}
This work was supported by the European Project GRIFFIN ERC Advanced Grant 2017, Action 788247. Partial funding was obtained from the Plan Estatal de Investigación Científica y Técnica y de Innovación of the Ministerio de Universidades del Gobierno de España (FPU19/04692).

\bibliographystyle{IEEEtran}
\bibliography{auro2021}

\end{document}